\documentclass{article}

\usepackage{geometry}

\usepackage{hyperref}
\usepackage{algorithm}
\usepackage{algpseudocode}
\usepackage{graphicx}
\usepackage{subcaption}
\usepackage{xcolor, colortbl}
\usepackage{booktabs}
\usepackage{amsmath}
\usepackage{amsfonts}
\usepackage{mathtools}   
\usepackage{verbatim}
\usepackage{marginnote}

\usepackage{fancyhdr}

\let\[\equation
\let\]\endequation

\begin{document}
\newcommand{\norm}[1]{\left| #1 \right|_2}
\newcommand\relatedversion{}
\renewcommand\relatedversion{\thanks{The full version of the paper can be accessed at \protect\url{https://arxiv.org/abs/1902.09310}}} 

\title{COMBOOD: A Semiparametric Approach for Detecting Out-of-distribution Data for Image Classification}
\author{%
Magesh Rajasekaran
\thanks{
  Louisiana State University\\
  Baton Rouge, LA 70803\\
  \texttt{mrajas1@lsu.edu} \\}
  \and
 Md Saiful Islam Sajol\textsuperscript{*}
  \and
  Frej Berglind\textsuperscript{*}
 \and
 Supratik Mukhopadhyay\textsuperscript{*}
  \and 
Kamalika Das
\thanks{Intuit Inc.\\
  2700 Coast Ave \\
Mountain View CA 94043\\
\texttt{kamalika\_das@intuit.com}
}
}
\date{}


\setlength{\headheight}{14pt} 
\pagestyle{fancy}
\fancyhf{}
\fancyhead[C]{\footnotesize\itshape First Published in Proceedings of the 2024 SIAM International Conference on Data Mining (SDM24)   

DOI: https://doi.org/10.1137/1.9781611978032.74}
\renewcommand{\headrulewidth}{0.4pt}


\maketitle

\thispagestyle{fancy}






\begin{abstract} Identifying out-of-distribution (OOD) data at inference time is crucial for many  machine learning applications, especially for automation. We present a novel unsupervised semi-parametric framework  \texttt{COMBOOD} for  OOD detection  with respect to  image recognition. Our framework combines signals from two distance metrics,  nearest-neighbor  and  Mahalanobis, to derive a confidence score for an inference point to be  out-of-distribution.  The former provides a non-parametric approach to OOD detection. 
The latter provides a parametric, simple, yet effective method for detecting OOD data points, especially, in the \emph{far OOD} scenario, where the inference point is far apart from the training data set in the embedding space. However, its performance is not satisfactory in the \emph{near OOD} scenarios that arise in practical situations.  Our \texttt{COMBOOD} framework combines the two signals in a semi-parametric setting to provide a confidence score that is accurate both for the near-OOD and far-OOD scenarios.  We show experimental results with the \texttt{COMBOOD} framework for different types of feature extraction strategies. We demonstrate experimentally that \texttt{COMBOOD} outperforms state-of-the-art OOD detection methods on  the OpenOOD  (both version 1 and  most recent version 1.5) benchmark datasets (for both far-OOD and near-OOD) as well as on the documents dataset  in terms of  accuracy. 

On a majority of the  benchmark datasets, the improvements in accuracy resulting from the \texttt{COMBOOD} framework  are statistically significant. \texttt{COMBOOD} 
scales linearly with the size of the embedding space, making it ideal for many real-life applications.\end{abstract}

~
\section{Background and Motivation} 
Deep neural networks are known to be opaque in their decision-making process \cite{jeyakumar2020can}. This becomes problematic when autonomous decisions need to be made on inputs whose salient characteristics can possibly be  different from what the neural network has been trained to identify. Such situations frequently arise when dealing with out-of-distribution (OOD) data \cite{yang2021generalized}, i.e., test data at inference time that does not come from the training distribution. For example, a neural network that has been trained to classify images of horses and giraffes, when provided with an image of an elephant, will classify it either as a horse or a giraffe, rather than determining that it does not belong to either of the classes.   Indeed, the fundamental assumption in supervised machine learning  that both the training and the test data come from the same probability distribution \cite{bishop1995neural,bishop2006pattern}, is violated in many real-world applications like automatic text extraction from documents, medical diagnosis, autonomous driving, etc. In such situations, neural networks can make erroneous decisions rather than issuing a warning that they have encountered OOD data on which their decisions cannot be trusted.   When document text extraction models are deployed in real-life applications, they may encounter a crumpled document that is damaged beyond recognition. Instead of informing the users of the situation, the model will attempt to extract a (possibly incorrect) set of texts from it. In safety-critical decision-making such as medicine, finance, autonomous driving, and mission-critical geospatial applications \cite{bas2015}, for machine learning models to be trusted, it is important for the model to identify when to abstain or require human intervention \cite{gunning2019xai}. 

It is well known \cite{baseline,clune} that the ``class probabilities" output by the softmax layer of a neural network are only weakly correlated with  how confident the model should be about the prediction, even in the relatively simple case of distinguishing Gaussian noise from in-distribution data \cite{baseline}. Measuring uncertainty associated with the decisions of a neural network during inference time is an active area of research \cite{ovadia2019can}. For uncertainties stemming from encountering OOD data at test time, various supervised and unsupervised methods have been proposed in the literature \cite{lee2018simple},\cite{sastry2020detecting}, \cite{odin}, \cite{devries2018learning}, \cite{sun2021react}, \cite{yang2022openood}, that provide varying degree of accuracy and come with varying degrees of computational overhead. Since convolutional layers have linear complexity, OOD detection algorithms for convolutional networks should ideally have  linear complexity in order to be useful and scalable in real-life applications. 

In recent times, effective post-hoc OOD detection methods based on distance metrics have been proposed \cite{lee2018simple,ren2021simple,sun2022knnood}. Among these,   Mahalanobis distance \cite{lee2018simple} provides a simple, effective method for identifying OOD data points, especially, in the \emph{far OOD} scenario, where the inference point is far apart from the training data set in the embedding space. However, its performance is not satisfactory \cite{ren2021simple} in the \emph{near OOD} scenarios that arise in practical situations. Sun et.al. \cite{sun2022knnood} proposed an OOD detection approach based on non-parametric nearest-neighbor distance that proved to be more accurate and computationally efficient than a Mahalanobis distance-based approach. However, its performances \cite{yang2022openood} on many far and near OOD detection tasks when trained on benchmark datasets like CIFAR 100, ImageNet, etc.,  leave  room for improvement. Developing OOD detection algorithms with improved performances compared to these baselines is important especially when neural network models are used to automate critical tasks such as extracting information from handwritten tax documents or physician prescriptions. 

We present a novel unsupervised semi-parametric framework  \texttt{COMBOOD} for  OOD detection  with respect to  image classification. Our framework combines signals from two distance metrics,  nearest-neighbor  and  Mahalanobis, to derive a confidence score for an inference point to be  out-of-distribution.  The former  provides a non-parametric approach to OOD detection. 
The latter provides a parametric, simple, yet effective method for detecting OOD data points, especially, in the \emph{far OOD} scenario, where the inference point is far apart from the training data set in the embedding space. However, its performance is not satisfactory in the \emph{near OOD} scenarios that arise in practical situations. 


Our \texttt{COMBOOD} framework combines the two signals in a semi-parametric setting to provide a confidence score that is accurate for both the \emph{near OOD} and \emph{far OOD} scenarios \cite{yang2022openood} (i.e., it has reduced area under the intersection of the distributions of the distances of the in-distribution and OOD datasets)> 
We show experimental results with the \texttt{COMBOOD} framework for different types of feature extraction strategies.

  Our first feature extraction strategy relies  on the computation of global extrema of the input features. 
  


  Our second feature extraction strategy extracts embeddings from the penultimate layer of a pretrained network. The penultimate layer embeddings   have the   representation of the image at the highest level of abstraction. We normalize the penultimate layer embeddings  by dividing them by their $L_2$ norms. While this feature extraction strategy works extremely well with the nonparametric nearest-neighbor distance-based approach \cite{sun2022knnood}, its combination with the parametric Mahalanobis distance-based approach significantly slows down the latter in terms of computation time.

\paragraph{Contributions:} This paper makes the following contributions
\begin{itemize}
\item  We present a novel unsupervised semi-parametric framework  \texttt{COMBOOD} for  OOD detection  in the context of  image classification. Our framework combines signals from two distance-based methods, nearest neighbor (non-parametric)  and Mahalanobis (parametric), to derive a confidence score for an inference point to be  out-of-distribution.  

   
   We demonstrate experimentally that \texttt{COMBOOD} outperforms state-of-the-art OOD detection methods on  the OpenOOD benchmark datasets (for far-OOD and near-OOD for both OpenOOD version 1 \cite{yang2022openood} and the recent version 1.5 \cite{zhang2023openood}) as well as on the documents dataset \cite{larson-2022-rvl-cdip-ood}  in  accuracy.
   For a majority of the  benchmark datasets, the improvements in accuracy resulting from \texttt{COMBOOD} with respect to the state-of-the-art are statistically significant.
   
\item  We show experimental results with the \texttt{COMBOOD} framework for different types of feature extraction strategies.

\item \texttt{COMBOOD} 
scales linearly with the size of the embedding space, making it ideal for many real-life applications.
\end{itemize}

\section{Methods and Technical Solutions} 
In this section, we formally define the problem statement and discuss the theoretical background of our proposed method before describing the \texttt{COMBOOD} framework.

\def\t{\tilde}
\subsection{Problem Formulation}
A neural network based image classifier is a function  $f_\theta:{\cal I}\rightarrow Y $, where $\theta$ represents the weights of the network,  ${\cal I}$ is the input domain defined by a set of images and $Y=\{0,1,\ldots, K-1\}$, where $K$ is the number of classes in the input distribution. In the traditional image classification setting, the trained model $f_\theta$ outputs a decision $\hat{y} \in Y$ for every test image $\t{X}$. However, if $\t{X}$ is out of distribution with respect to the input distribution of ${\cal I}$, we would want the model $f_\theta$ to return, along with the class label $\hat{y}$, a score akin to model confidence that can indicate whether $\t{X}$ is out of distribution with respect to the marginal probability distribution $p_{{\cal I}}$ for the joint distribution $p({\cal I},Y)$. In other words, we want to learn a function $g$ that satisfies the following condition
\[
  g(\t{X}) =
  \begin{cases*}
                                   1 & \text{if $\t{X} \sim p_{{\cal I}}$} \\ 
                                   0 & \text{otherwise}
  \end{cases*}.
\]





\subsection{Feature Extraction Strategies} \label{feature}
We have experimented with two different feature extraction strategies. The first is a novel technique that involves computing global extrema of the input features. Using this feature extraction strategy, an unsupervised  Mahalanobis distance-based approach provides accurate results on many far OOD detection tasks while significantly improving on computational efficiency compared to \cite{lee2018simple}. Yet, like \cite{lee2018simple}, its performance on near OOD detection tasks  remains far from satisfactory. The second strategy extracts embeddings from the penultimate layer of the pretrained network. Its use with the parametric Mahalanobis distance-based approach significantly slows the latter \cite{lee2018simple} down without any improvement in performance. However, its use with the nearest-neighbor distance-based approach  provides an OOD detection method that outperforms the Mahalanobis distance-based approach both in terms of accuracy and computational efficiency, for both far and near OOD scenarios. 
\subsubsection{Extracting Global Extrema of Input Features} \label{extrema}
The essence of this feature extraction strategy is in the nature of the distribution of the extreme values of the data as it enters the activation layers in a neural network. We have observed that the per activation layer distribution of maximum or minimum value is consistently unimodal on in-distribution data.

Based on this observation, we compute the global maximum and minimum of the input images before each activation layer in the architecture (see Function 1). To improve the symmetry of this distribution, we apply a Yeo-Johnson transform \cite{yeo-johnson} fitted on the extreme values of the training data. The output of this transform is standardized to zero mean and unit variance. This computation does not make use of class labels or image topology, yet it is highly effective for OOD detection as we show below. 

\subsubsection{Extracting Embeddings from the Penultimate Layer} \label{penultimate}
This strategy extracts embeddings from the penultimate layer of a pretrained network. The penultimate layer embeddings have the  information at the highest level of abstraction  compared to other layers in the network. The extracted penultimate embeddings are then normalized as follows. Let ${\cal I}$ be the set of all ${\cal M} \times {\cal M}$ images with pixel values from $\{0, \ldots, L-1\}$. 
Let $\Theta: {\cal I} \rightarrow {\cal R}^n$ be a feature encoder. Then, for an input $J \in \cal{I}$, the normalized penultimate layer embedding is given by $\Theta(J)/\norm{\Theta(J)}$, where $\norm{.}$ denotes the $L_2$ norm.






\begin{algorithm}[h]
    \caption{\textbf{Function 1:} ExtremeValueExtraction}
        \begin{algorithmic}[1]
            \Require Data set $X_t$
            \For{\texttt{each image $X_{t,i}$ in training set} $X_t$}
                \State $X_{min,i}$, $X_{max,i}$ = \texttt{MinMax}($X_{t,i}$)
            \EndFor
            \State $X_{min}$ = $[X_{min,1}, X_{min,2}, X_{min,3}, X_{min,4},\dots ]$
            \State $X_{max}$ = $[X_{max,1}, X_{max,2}, X_{max,3}, X_{max,4},\dots ]$
            \State $X_{j}^{P}$, $T_{j}^{P}$ = \texttt{fit.PowerTransform}($X_{min,j}$, $X_{max,j}$)
            \State \textbf{return} $x_{j}^{P}$
\label{fun:fe}
\end{algorithmic}
\end{algorithm}

\section{Algorithms for OOD Detection}
We first start with a parametric algorithm for OOD detection based on regularized Mahalanobis distance. 
\subsection{Regularized Mahalanobis Distance\label{md}}
Let $\mu$ and $M$ be the mean and covariance of the features extracted on the training set. We modify $M$ by adding a constant to the diagonal:

\[
M' = M + C \cdot I,
\]
where $C \in [0, \infty)$ is a regularization constant and $I$ is the identity matrix. 

We assume that  the feature values are standardized, so the diagonal of $M$ is 1 and we can expect $C$ to have a consistent effect on each instance. 

The Mahalanobis distance $D_M$ is computed using the following equation where $x$ is a feature vector extracted from test image using one of the feature extraction strategies described above 
\[
D_M(x) = \sqrt{(x - \mu)^\mathsf{T} (M + C \cdot I)^{-1} (x - \mu)}.
\]
The value $C=0$, corresponds to the standard Mahalanobis distance, that represents  the likelihood of a multivariate normal distribution. If $C$ is large, the regularized covariance matrix $M'$ and the corresponding inverse covariance matrix $(M')^{-1}$ are dominated by the constant on the diagonal and $D_M$ is approximately proportional to the $L^2$ distance. This corresponds to the likelihood of a normal distribution where each of the extreme values are independent. The regularizer $C$ allows us to control the definition of in-distribution data. Algorithm 1 presents the pseudo-code of the regularized Mahalanobis distance-based  parametric algorithm. 




\begin{algorithm}
\caption{\textbf{Algorithm 1:} Regularized Mahalanobis Distance}\label{alg:m}
\begin{algorithmic}
\Require Training image and label set $<X_t, y_t>$ , Pretrained classifier $f_\theta$, Regularization $C$
\Ensure Mean $\mu$ and Covariance $K$ of OOD features
\State \texttt{Predict} $\hat{y}_t$ = $f_\theta(X_t)$
\State $X_{t}$ = $X_{t, \hat{y}_t==y_t}$
\For{each input $X'_{t,j}$ to activation layer $j$ in $f_\theta$ }
\State $X_{j}^{P}$ = \texttt{FeatureExtraction}($X'_{t,j}$)
\EndFor
\State $\mu$ = \texttt{Mean}($X_{j}^{P}$)
\State $M$ = \texttt{Covariance}($X_{j}^{P}$)
\State $D_M = \sqrt{(X_{j}^{P} - \mu)^\mathsf{T} (M + C \cdot I)^{-1} (X_{j}^{P} - \mu)}$
\State \textbf{return} $D_M$.
\end{algorithmic}
\end{algorithm}
  The regularized Mahalanobis Algorithm 1 can use either of the feature extraction strategies (global extrema of input features or penultimate layer embeddings) as its \texttt{FeatureExtraction} method.  It operates without apriori knowledge of the OOD data. It is completely unsupervised. 
  
  When using the global extrema of input features, it only computes some statistics on the extreme value distribution of the training data and uses the notion of  distance from those statistics to identify OOD data points, and  
  has minimal computational overhead. The only features that are computed on the training data are mean and covariance on the  feature set which is low dimensional with only $2\times r$ features, where $r$ is the number of activation layers in the network.

  
  In experiments, this  method has proved to be  extremely efficient and accurate and produces state-of-the-art results on all benchmark data sets for which the extreme value distribution of the training data is compact and well separated from the OOD data. Yet, like \cite{lee2018simple}, its performance on near OOD detection tasks  remains far from satisfactory (see the Experiments section below).
  In addition, when this method is used with embeddings from the penultimate layer as features,  there is a significant increase in computational time without any improvement in performance.

  \subsection{Nearest Neighbor Distance}
This algorithm uses  non-parametric density estimation using nearest neighbor, i.e., we compute the k-nearest neighbor between the  test image and the training images in the embedding space \cite{sun2022knnood}.  We then classify it as OOD or ID based on a threshold value typically picked so that ID data are classified correctly \cite{sun2022knnood}. The algorithm  ingests the  features extracted from the pretrained network; we experimented with both types of features: global extrema of the features extracted from the layers of the pretrained network as well as the embeddings extracted from the penultimate layer of the network. It surpasses the parametric regularized Mahalanobis distance-based algorithm (with both feature extraction strategies) in terms of both accuracy and inference time on all the OpenOOD benchmark datasets (both version 1 \cite{yang2022openood} and more recent version 1.5 \cite{zhang2023openood}) as well as on the documents dataset. However, there is still  room for improvement, especially in applications where critical tasks are automated  such as extracting information from handwritten tax documents or physician prescriptions. 
 \section{COMBOOD: A Semiparametric Framework} 
  To improve upon the above two baseline distance-based algorithms above, we propose a semiparametric framework  \texttt{COMBOOD}   that combines  the parametric regularized Mahalanobis distance-based approach with the non-parametric nearest neighbor distance-based approach. 




 \texttt{COMBOOD} combines the signals obtained from both the  nearest-neighbor distance  and the regularized Mahalanobis distance  using a log distribution of respective distances. Let the nearest neighbor distance be $kd$ and the  regularized Mahalanobis distance be $md$. \texttt{COMBOOD} computes  confidence scores, $kc$ and $mc$ respectively,  from the log distributions of  the nearest neighbor distance and the regularized Mahalanobis distance using the following equations, where $k$ is the parameter used in the nearest neighbor algorithm. $n$ is the dimensionality of the feature space,  $M'$ is the modified  covariance Matrix in the regularized Mahalanobis distance as defined above, and $c$ is a constant. Finally, \texttt{COMBOOD} computes the final score by adding the confidence scores $kc$ and $mc$. We refrain from using weighted addition which would have required the user to choose ad-hoc weights. In this way, \texttt{COMBOOD} combines the evidences (i.e., priors) provided by the parametric regularized Mahalanobis distance-based approach and the nonparametric nearest neighbor distance-based approach to provide support ($score$) for classifying a test image as OOD. 



 \[
kc = - \sqrt{n} * \log(kd)
\]

\[
mc = \log(\exp{(-md^2 / 2)}/\sqrt{det(2\pi M')})
\]

\[
score = kc + mc
\]



 We experimentally show that \texttt{COMBOOD} performs best when the nonparametric and the parametric components use different feature extraction strategies, penultimate layer embeddings for the former and global extrema of the features for the latter. 
  
  This is due to two important reasons. The first is that, as the data passes through each layer of the pre-trained model, we lose some information.  While extracting embeddings  from the penultimate layer,  we only get the highest level of abstraction of the data.  In computing the global extrema of the input features,  we get information from all the layers. Furthermore, while normalizing the feature vectors from the penultimate layer, we lose information  on the magnitude of the feature vector. However, when extracting the  global extrema of features in every layer, we get back some information about the magnitude of the data. Thus these two feature extraction strategies  complement each other and help in both \texttt{far}-OOD and \texttt{near}-OOD scenarios.


\subsection{Design Choices in Models} \label{design}

\begin{figure}[htbp]
\centering
\includegraphics[width=.5\textwidth]{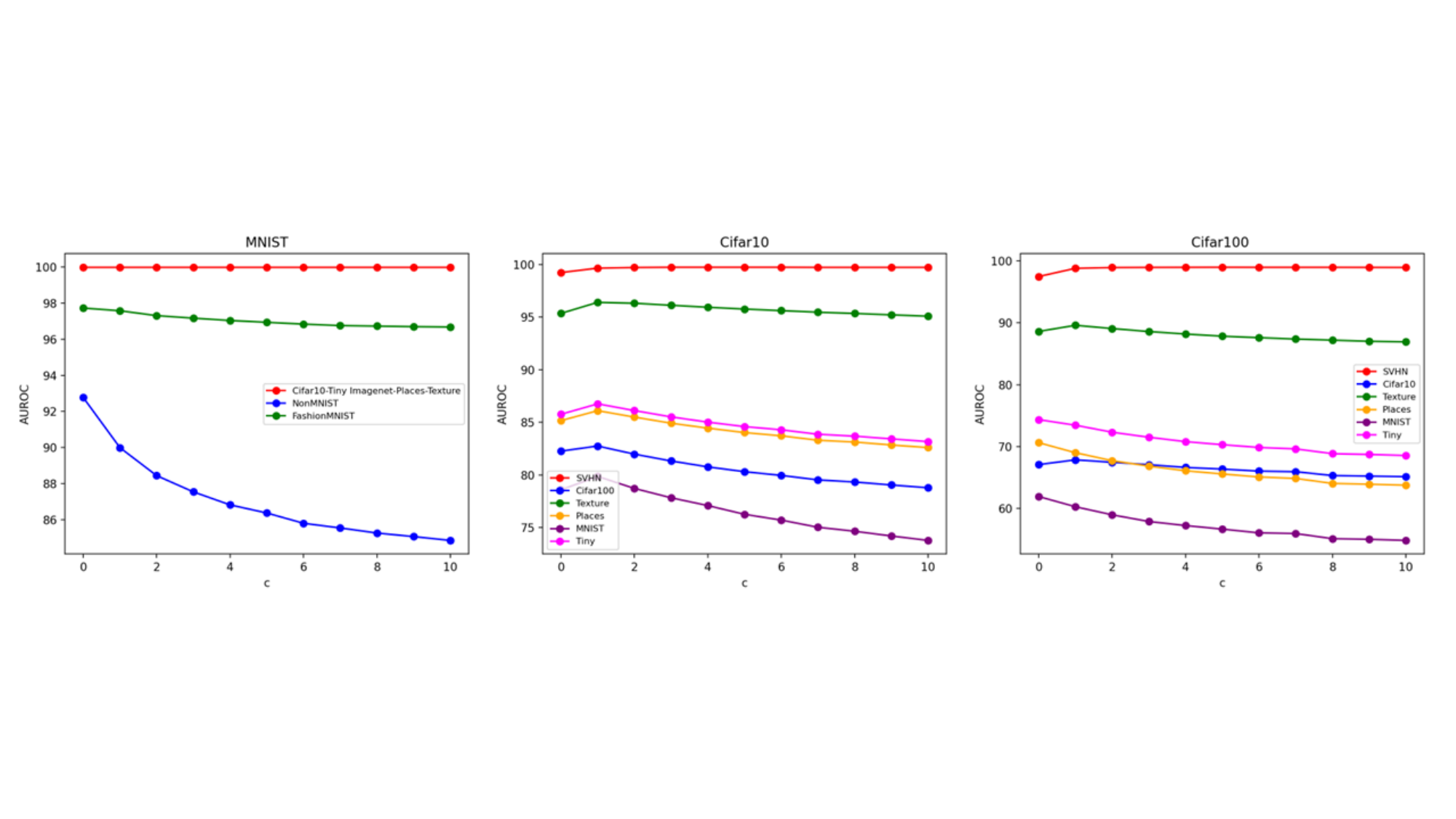}
\caption{AUROC-score of \texttt{Regularized Mahalanobis} for various regularization values. The score was computed on the test set of the different dataset such as CIFAR10, CIFAR100 and MNIST.}\label{fig:regularization}
\label{Mahala Tuning}
\end{figure}

\textbf{Regularization}
The $\ell_2$ regularization in the regularized Mahalanobis-based method  cannot be learned through cross validation. So we experimented with a wide range of values for $C$. We found that a regularization $C=1$ worked well across all architectures and data sets. Figure \ref{fig:regularization} shows a comparison of detection performance with $C$ ranging from 0 to $\infty$, where $\infty$ corresponds to $L_2$-distance instead of Mahalanobis distance.

\section{Empirical Evaluation} \label{experiment} 

In this section, we report the results obtained from running parametric regularized Mahalanobis distance-based approach and the semiparametric \texttt{COMBOOD} framework  on OpenOOD benchmark datasets \cite{yang2022openood,zhang2023openood} and the documents dataset \cite{larson-2022-rvl-cdip-ood} on a number of performance metrics. The code is available at \cite{code1}. We track performance with respect to the quality of the results as well as the efficiency of the methods. For quality of results, we measure performance based on the following metrics:

\begin{itemize}
    \item \textbf{AUROC:} The area under plot of true positive rate (TPR) versus false positive rate (FPR). A random detector has AUROC = 50\% and an ideal detector has AUROC = 100\%.
    \item \textbf{Inference Time:} We calculate the per-image inference time of different datasets in milliseconds averaged across test images.
    \item \textbf{AUPR:} curve represents the relation between precision and recall.
\end{itemize}
\subsection{Results}
We have evaluated the parametric regularized Mahalanobis distance-based approach, the nonparametric nearest-neighbor-distance-based approach, and the semiparametric \texttt{COMBOOD} framework on ResNet18 for Cifar-10, Cifar-100 and on LeNet for MNIST, and on ResNet50 for Imagenet1-k (as benchmarked by the OpenOOD framework) using the same pretrained models as used by the OpenOOD framework \cite{zhang2023openood} for uniform comparison. In Tables \ref{Table:ResNet18} and \ref{Table:ResNet50} (the tables are presented in the same format as in the OpenOOD framework \cite{yang2022openood,zhang2023openood}), we compare our algorithms with the  ODIN \cite{odin,zhang2023openood}, Mahalanobis  as in\cite{lee2018simple,zhang2023openood}, Gram matrix \cite{sastry2020detecting,zhang2023openood} and \texttt{KNN} \cite{zhang2023openood}. \texttt{COMBOOD} outperforms  ODIN and Mahalanobis \cite{lee2018simple}  in all experiments for both near and far OOD detection scenarios. And even though Gram is far more computationally expensive, \texttt{COMBOOD} outperforms it in all far and near OOD detection scenarios. \texttt{COMBOOD} outperforms \texttt{KNN} \cite{zhang2023openood} in most of the \texttt{near}-OOD and \texttt{far}-OOD scenarios. 
\definecolor{celadon}{rgb}{0.67, 0.88, 0.69}
\definecolor{ao}{rgb}{0.0, 0.0, 1.0}
\definecolor{amaranth}{rgb}{0.9, 0.17, 0.31} 

\begin{table*}[t]
\resizebox{\textwidth}{!}{
\begin{tabular}{llccc}
\toprule
     &      &     AUROC                                      & AUPR \\
In Dist & Out Dist &                                                  &                                                              \\
\midrule
  CIFAR-10     & \color{ao}{CIFAR100} &                        \color{ao}{77.68/89.76/66.3/82.78/89.73/\textbf{90.08}} &          \color{ao}{73.24/59.04/63.74/79.71/88.22/\textbf{89.95}} \\

     & \color{ao}{Tiny} &              \color{ao}{77.33/58.11/66.79/86.77/91.56/\textbf{92.18}} & \color{ao}{70.07/54.72/63.28/85.6/88.77/\textbf{92.97}} \\
     
     & \color{ao}{\textbf{NearOOD}} &                              \color{ao}{77.51/58.57/66.54/84.77/90.64/\textbf{91.13}} & \color{ao}{71.66/56.18/63.51/85.18/88.50/\textbf{91.46}}\\
     
     & \color{amaranth}{MNIST} &                            \color{amaranth}{90.91/77.59/99.52/79.81/\textbf{94.26}/91.28} & \color{amaranth}{64.74/43.97/\textbf{99.24}/34.8/98.95/74.49} \\
     
     & \color{amaranth}{SVHN} &                     \color{amaranth}{73.32/79.28/95.78/99.67/92.67/\textbf{99.06}} & \color{amaranth}{42.13/55.46/91.00/99.05/96.25/\textbf{97.79}} \\
     
     & \color{amaranth}{Texture} &                              \color{amaranth}{80.7/57.72/95.42/96.41/93.16/\textbf{95.98}} &   \color{amaranth}{82.25/62.75/96.69/97.58/87.26/\textbf{97.66}}        \\
     
     & \color{amaranth}{Places365} &                              \color{amaranth}{82.55/55.44/64.40/86.13/91.77/\textbf{92.27}} &    \color{amaranth}{50.27/24.06/31.81/60.1/97.24/\textbf{80.46} }      \\
     
     & \color{amaranth}{\textbf{FarOOD}} &                              \color{amaranth}{81.87/67.51/88.78/90.50/92.96/\textbf{94.64}} &   \color{amaranth}{59.85/46.56/79.68/ 72.88/94.93/\textbf{87.6}}         \\
     
CIFAR-100     & \color{ao}{CIFAR10}  &                              \color{ao}{78.18/54.33/46.99/68.00/\textbf{77.02}/75.39} &         \color{ao}{\textbf{79.12}/53.50/46.02/63.19/75.36/71.65}  \\

     & \color{ao}{Tiny} &                      \color{ao}{81.39/56.38/55.73/73.65/\textbf{83.35}/82.14} &  \color{ao}{85.30/61.29/59.23/75.96/\textbf{73.99}/86.1}         \\
     
     & \color{ao}{\textbf{NearOOD}} &                              \color{ao}{79.79/55.35/51.36/70.82/\textbf{80.18}/78.76} &     \color{ao}{82.21/57.39/52.63/69.57/\textbf{74.67}/78.87}       \\
     
     & \color{amaranth}{MNIST} &                              \color{amaranth}{83.71/71.86/\textbf{97.17}/60.36/82.36/77.74} &   \color{amaranth}{62.02/40.07/\textbf{94.12}/17.07/96.19/55.64}          \\
     
     & \color{amaranth}{SVHN} &                               \color{amaranth}{71.08/95.68/56.69/\textbf{98.78}/84.15/97.69} &   \color{amaranth} {52.36/90.63/28.53/\textbf{96.97}/92.79/95.2}       \\
     
     & \color{amaranth}{Texture} &                     \color{amaranth}{79.39/74.80/68.00/89.59/83.66/\textbf{90.19}} &    \color{amaranth}{86.67/78.41/67.66/93.09/74.81/\textbf{93.75}}        \\
     
     & \color{amaranth}{Places365} &                      \color{amaranth}{\textbf{79.83}/48.47/58.73/69.16/79.43/77.85} &    \color{amaranth}{\textbf{60.85}/21.48/28.11/37.00/91.98/55.09 }       \\
     
     & \color{amaranth}{\textbf{FarOOD}} &                    \color{amaranth}{78.50/72.70/70.14/79.47/82.40/\textbf{85.86}} &   \color{amaranth}{65.47/57.65/54.60/61.03/88.94/\textbf{74.92} }        \\
     
\bottomrule
\end{tabular}
}
\caption{\textbf{ResNet18} detection scores (benchmark results from \cite{zhang2023openood}) for ODIN/MDS/Gram/Reg. Mahalanobis/KNN/COMBOOD. Results for other algorithms are from  Sastry and Oore \cite{sastry2020detecting} and KNN \cite{yang2022openood}. Near OOD dataset results are in blue while far OOD ones are in red. The best results are in bold.} \label{Table:ResNet18}
\end{table*}

\begin{table*}[h] 
\resizebox{\textwidth}{!}{
\begin{tabular}{llccc}
\toprule
     &      &                                         AUROC &  AUPR        \\
In Dist & Out Dist &                                                  &         \\
\midrule
MNIST     & \color{ao}{NotMNIST} &             \color{ao}{89.85/50.19/\textbf{99.59}/89.98/97.37/98.20} &      \color{ao}{80.76/32.84/\textbf{99.16}/82.67/93.85/96.62 }   \\

     & \color{ao}{FashionMNIST} &             \color{ao}{94.91/97.61/96.41/97.57/95.68/\textbf{98.71}} & \color{ao}{94.46/97.54/96.37/97.26/95.10/\textbf{98.63}}  \\
     
     & \color{ao}{\textbf{NearOOD}} &                               \color{ao}{92.38/73.90/98.00/93.78/96.52/\textbf{98.45}} &   \color{ao}{87.61/65.19/\textbf{97.76}/89.96/94.47/97.62}         \\
     
     & \color{amaranth}{Texture} &                       \color{amaranth}{98.99/99.72/98.54/\textbf{99.98}/94.01/\textbf{99.98}} &   \color{amaranth}{99.25/99.72/99.12/\textbf{99.99}/95.74/\textbf{99.99}}        \\
     
     & \color{amaranth}{CIFAR10} &                     \color{amaranth}{99.11/99.75/98.45/99.99/97.70/\textbf{100.0}} &  \color{amaranth}{99.09/99.74/98.61/99.99/97.58/\textbf{100.0}}          \\
     
     & \color{amaranth}{Tiny} &                             \color{amaranth}{99.22/99.75/97.71/\textbf{99.99}/97.40/\textbf{99.99}} & \color{amaranth}{99.20/99.77/97.89/\textbf{99.99}/97.08/\textbf{99.99}}           \\
     
     & \color{amaranth}{Places365} &                              \color{amaranth}{98.76/99.78/97.78/99.99/97.52/\textbf{100.0}} & \color{amaranth}{96.29/99.32/94.80/99.97/92.47/\textbf{99.99} }         \\
     
     & \color{amaranth}{\textbf{FarOOD}} &              \color{amaranth}{99.02/99.75/98.12/\textbf{99.99}/96.66/\textbf{99.99}} &  \color{amaranth}{98.46/99.64/97.60/\textbf{99.99}/95.72/99.99 }         \\
     
\bottomrule
\end{tabular}
}
\caption{\textbf{LeNet} detection scores (benchmark scores are from \cite{zhang2023openood}) for ODIN/MDS/Gram/Reg. Mahalanobis/KNN/COMBOOD. Results for other algorithms are from  Sastry and Oore \cite{sastry2020detecting}. Near OOD dataset results are in blue while far OOD ones are in red. The best results are in bold.} \label{Table:LeNet}
\end{table*}

\begin{table*}[!]
\resizebox{\textwidth}{!}{%
\begin{tabular}{llccc}
\toprule
     &      &                                                  AUROC & AUPR \\
In Dist & Out Dist &                                                  &\\
\midrule
ImageNet     & \color{ao}{Species} &                             \color{ao}{71.59/67.26/60.20/64.53/76.37/\textbf{77.71}} &  \color{ao}{91.38/89.90/87.55/88.54/93.04/\textbf{93.36}} \\

     & \color{ao}{iNaturalist} &                              \color{ao}{\textbf{91.19}/78.39/67.84/71.63/85.04/87.13} &  \color{ao}{\textbf{97.72}/4.07/90.94/91.58/96.35/96.3} \\
     
     & \color{ao}{OpenImage-O} &                              \color{ao}{\textbf{88.35}/71.88/70.74/64.31/86.42/86.6} &  \color{ao}{95.28/86.84/86.39/82.91/94.85/94.84} \\
     
     & \color{ao}{ImageNet-O} &                               \color{ao}{41.45/55.82/74.31/50.63/\textbf{75.39}/74.08} & \color{ao}{95.41/96.36/98.19/\textbf{95.87}/98.58/98.48}  \\
     
     & \color{ao}{\textbf{NearOOD}} &                              \color{ao}{73.15/68.34/68.27/62.77/80.81/\textbf{81.38}} &  \color{ao}{94.95/91.79/90.77/89.72/95.70/\textbf{95.74}} \\
     
     & \color{amaranth}{Texture} &                              \color{amaranth}{89.17/82.72/90.23/82.8/96.18/\textbf{97.0}} &  \color{amaranth}{98.25/96.68/97.35/97.13/99.48/\textbf{99.59}} \\
     
     & \color{amaranth}{MNIST} &                               \color{amaranth}{99.67/95.66/97.69/93.56/99.85/\textbf{99.88}} &  \color{amaranth}{99.94/99.16/99.60/98.84/99.97/\textbf{99.98}} \\
     
     & \color{amaranth}{\textbf{FarOOD}} &                            \color{amaranth}{94.42/89.19/93.96/88.18/98.01/\textbf{98.44}} &  \color{amaranth}{99.09/97.92/98.47/97.98/99.73/\textbf{99.78}} \\
     
\bottomrule
\end{tabular}
}
\caption{\textbf{ResNet50} detection scores (benchmark scores are from \cite{yang2022openood}) for ODIN/MDS/Gram/Reg. Mahalanobis/KNN/COMBOOD. Results for other algorithms are from Sastry and Oore \cite{sastry2020detecting}. Near OOD dataset results are in blue while far OOD ones are in red. The best results are in bold.} \label{Table:ResNet50}
\end{table*}

\begin{table*}[!] 
\centering
\resizebox{11.5cm}{!}{%
\begin{tabular}{llccc}
\toprule
  In Dist   &   Out Dist   &    AUROC                                                 & AUPR \\
\midrule
Document    & RVL\_CDIP\_N &                                90.74/97.35/\textbf{98.52} & 99.71/99.93/\textbf{99.96}\\
     & RVL\_CDIP\_O &                             90.38/97.78/\textbf{98.72} & 99.15/99.83/\textbf{99.9} \\
\bottomrule
\end{tabular}
}
\caption{\textbf{ResNet18} detection scores for Reg. Mahalanobis/KNN/COMBOOD. Datasets are derived from \cite{larson-2022-rvl-cdip-ood}}
\label{Table:Document12}
\end{table*}

\subsection{Inference Time} \label{inference}
In real-world applications of OOD-detection, the inference time is often crucial, yet this has largely been neglected in research. We measured the inference time of a ResNet18 on Cifar10 and Cifar100 for the baseline \cite{baseline}, Mahalanobis \cite{lee2018simple}, Gram \cite{sastry2020detecting}, \texttt{KNN} \cite{zhang2023openood} and \texttt{COMBOOD}. 

For \texttt{COMBOOD}, We chose to show the inference time of the test images by averaging the test set following  the per-image-inference time of \texttt{KNN} benchmarking scheme used by the OpenOOD framework \cite{yang2022openood}. 
The results are shown in Table \ref{Table: Time ResNet}. The two components of   \texttt{COMBOOD} can be processed simultaneously to calculate the distances for the nonparametric and the parametric components  and the resulting priors can be combined to get the final confidence score. Since the execution is parallelized, the inference time is calculated for both the component approaches and the maximum among them is recorded. All experiments were conducted on a single \textit{NVIDIA Tesla V100} GPU and one \textit{Intel Xeon Silver 4216} CPU with a batch size of 128, averaging over 10 runs.



\begin{table}[H] 
\resizebox{7cm}{!}{
\begin{tabular}{lr}
\toprule
Dataset & Inference Time (ms) \\
\midrule
Species    &       5.78    \\
OpenImage-O     &            5.33  \\
MNIST    &            5.17  \\
ImageNet-O &          6.3    \\
iNaturalist       &    5.05  \\
Texture        &          5.22  \\
\midrule
Average & 5.4 \\
\bottomrule
\end{tabular}

}
\caption{\label{Table: Time ResNet}
Average per-image-inference time measured in milliseconds for \texttt{COMBOOD} for various test datasets of ImageNet-1k on ResNet50.}
\end{table}


\subsection{Further Experiments}
We conducted further experiments on the \texttt{COMBOOD} framework using different feature extraction strategies as well as conducting a case study on a document dataset \cite{larson-2022-rvl-cdip-ood}.

\subsubsection{Case Study for Document Dataset}
We also evaluated the \texttt{COMBOOD} framework  against two of the important document datasets provided in  \cite{larson-2022-rvl-cdip-ood}. Our \texttt{COMBOOD} algorithm outperforms the state of the art, \texttt{KNN}, \cite{sun2022knnood} algorithm with a significant increase in AUROC to 1.5. We analyzed the document images which were correctly classified by our \texttt{COMBOOD} but missed by \texttt{KNN} algorithm to understand the importance of our semiparametric approach. We trained the document dataset with training set of 1000 images and used a test set of  500 images. We found that out of 500 test images \texttt{KNN} algorithm wrongly classified 79 images whereas our \texttt{COMBOOD} wrongly classified only 39 images, thus reducing the error rate from 15.8\% to 7.8\%. The result are provided in the Table \ref{Table:Document12}.

\subsection{Statistical Significance Testing}
To evaluate the statistical significance of the improvements in AUROC provided by the COMBOOD framework over the KNN (among the posthoc methods evaluated in \cite{zhang2023openood}, KNN performs the best) we performed the \texttt{McNemar test} \cite{McNemar1947}. The test showed that  the improvements provided by the   \texttt{COMBOOD} framework over the KNN \cite{zhang2023openood} are statistically significant for most of the benchmark datasets. 
\section{Related Work}
The last few years have seen a tremendous amount of work in OOD detection \cite{zhang2023openood,odin, sastry2020detecting,sun2021react,ren2019likelihood,odin,lee2018simple,ren2019likelihood,devries2018learning,liu2020energy,vyas2018out,yu2019unsupervised,lin2021mood,mohseni2020self,huang2021mos,kumar2021calibrated,hendrycks2019scaling}. However, as argued in \cite{tajwar2021no}, no single algorithm can be considered the state-of-the-art. In \cite{tajwar2021no}, the authors tested the algorithms in \cite{odin,baseline,lee2018simple} on three benchmark in-distribution datasets (CIFAR-10, CIFAR-100, SVHN) and seven benchmark OOD datasets under standardized conditions. They found inconsistent performance of these algorithms across all the datasets; in fact, there was no algorithm that was consistently outperforming others across all the datasets. 


Unlike algorithms like \cite{odin,lee2018simple,liu2020energy}, the \texttt{COMBOOD} framework does not  require access to the OOD data during training time. The openOOD framework \cite{yang2022openood,zhang2023openood} compared different OOD detection algorithms in a uniform benchmarking framework. In their evaluation, among post-hoc algorithms, KNN \cite{sun2022knnood} had the best performance both in terms of far and near OOD detection tasks. We have shown that  \texttt{COMBOOD} outperforms \texttt{KNN} \cite{zhang2023openood} in most of the \texttt{near}-OOD and \texttt{far}-OOD  detection scenarios. 


Mohseni et al. \cite{shifting_tranformation} compare different OOD detection algorithms in terms of inference time. All the algorithms considered, that are not allowed to train on OOD data,  have overhead above 100\%. Both the algorithms in \cite{sastry2020detecting,lee2018simple} involve matrix multiplication which has at least quadratic complexity. Since, the number of features within a neural network is large, matrix multiplication operations as used in \cite{sastry2020detecting,lee2018simple} are prohibitively expensive. In case of parametric regularized Mahalanobis distance-based approach with the global extrema of the features as input,  the extreme values  have dimensions twice the depth of the model (typically at most 1000); hence Mahalanobis distance can be computed efficiently. This explains why  this algorithm  outperforms \cite{sastry2020detecting} and \cite{lee2018simple} in terms of inference time by an order of magnitude. 


In \cite{siffer2017anomaly}, the authors used extreme-value theory for determining outliers in time-series data. Here, we have used a feature extraction strategy  that computes   global extrema of the input features. 



\section{Significance and Impact} 
We presented a novel unsupervised semi-parametric framework  \texttt{COMBOOD} for  OOD detection  with respect to  image classification. Our framework combines signals from two distance metrics,  nearest-neighbor  and  Mahalanobis, to derive a confidence score for an inference point to be  out-of-distribution.
We have demonstrated experimentally that \texttt{COMBOOD} outperforms state-of-the-art OOD detection methods on  the OpenOOD benchmark datasets (for both far-OOD and near-OOD \cite{yang2022openood}) as well as on the documents dataset  in terms of both efficiency and accuracy.
\subsection{Limitations of the Work}\label{limit}
The \texttt{COMBOOD} framework combines a nonparametric nearest neighbor distance-based method and a parametric regularized Mahalanobis distance-based method in a nonparametric setting for OOD detection. In particular, it combines the priors obtained by the two approaches. Currently, the way the priors are combined is not data-driven. In the future, we plan to investigate data-driven approaches for combining priors, in particular, considering the kurtosis of the ID and the OOD distributions. 
\subsection{Future Work}
In the future, we will study algorithms for detecting OOD for image classification both in the near and far OOD scenarios using the notions of layer and class applicability \cite{collier2018cactusnets} and extend the notions to generative models \cite{collier2021gap}.
\vspace{50pt}
\bibliographystyle{abbrv}
\bibliography{references}
\end{document}